\documentclass[11pt]{article}

\usepackage[preprint]{acl}

\usepackage{times}
\usepackage{latexsym}

\usepackage[T1]{fontenc}
\usepackage[utf8]{inputenc}

\usepackage{microtype}

\usepackage{inconsolata}

\usepackage{graphicx}
\usepackage{subcaption}
\usepackage{multirow}
\usepackage{listings}
\usepackage{tcolorbox}

\title{Fine-Tuning a KV Cache Concatenation-Aware Model or Recomputing KV Caches? Why Not Both?}

\author{Fumihiko Tachibana \\[0.5em]
  \\\And
  Daisuke Miyashita \\[0.5em]
  Kioxia Corporation \\\And
  Jun Deguchi \\[0.5em]
  \\}

\begin{document}
\maketitle
\begin{abstract}
In Retrieval-Augmented Generation (RAG) systems, a large number of retrieved chunks are concatenated to form the input context so that users can receive high-quality responses based on external knowledge.
As a result, the input context length increases substantially, leading to a larger prefill workload and, in turn, a longer time to first token (TTFT).
While previous works that reuse precomputed key-value (KV) caches effectively reduce TTFT for long-context inputs, it remains unclear whether response quality is preserved when the input context becomes very long.
In this paper, we propose a combined approach that (i) fine-tunes the model while taking KV cache concatenation into account and (ii) selectively recomputes a subset of the KV caches.
By applying both techniques, we demonstrate improved accuracy for long-context inputs.
Experiments on the RULER benchmark show that, for a 124k-token input, our method improves the RULER score by 9.7 point over the baseline that recomputes KV caches only.
Moreover, TTFT is reduced by 80\% compared with full attention.
\end{abstract}

\section{Introduction}

When a large language model (LLM) receives additional chunks retrieved by Retrieval-Augmented Generation (RAG) \cite{li2022surveyretrievalaugmentedtextgeneration, gao2024retrievalaugmentedgenerationlargelanguage}, the total input context length increases substantially.
Since prefill computation scales quadratically with input context length, computing the key-value (KV) caches incurs a large computational cost, which in turn increases the time to first token (TTFT).

One approach for reducing TTFT is to reuse precomputed KV caches for the context that is frequently reused.
First, prefix caching stores and reuses KV caches when the context begins with the same initial token sequence \cite{NEURIPS2024_724be447}.
Prefix caching does not degrade generation quality because the KV caches for the prefix context are independent of the subsequent text. 
However, this reuse is only valid when the input tokens exactly match those of the stored KV caches.
This constraint limits the reusability of KV caches.

Second, KV cache concatenation aims to reuse KV caches even when the context corresponding to the precomputed KV caches is not a prefix, thereby increasing their reusability.
However, because precomputed KV caches are generated without preceding chunks, reused KV caches for non-prefix tokens do not include cross-attention from preceding chunks, which introduces deviations in the KV caches compared with those computed with full attention.
When such a mismatch occurs, the attention outputs change, and generation quality degrades even if the positional embeddings of the KV caches are adjusted to match the current context positions \cite{ICLR2025_95306350}.

To mitigate this problem, several methods have been proposed.
CacheBlend \cite{yao2025cacheblend} and EPIC \cite{hu2025epic} selectively recompute portions of the KV caches based on certain criteria.
Other approaches, such as Block-attention \cite{ma2025blockattention}, TurboRAG \cite{lu-etal-2025-turborag}, and KVLink \cite{yang2025kvlink}, fine-tune the model under the assumption that there is no cross-attention between chunks, allowing multiple KV caches to be concatenated and reused.
Although prior techniques improve generation quality on several benchmarks, their response accuracy under long input contexts remains underexplored.
We show that accuracy consistently declines as context length increases, revealing that existing methods are inadequate for long-context settings.
Since the benefit of KV cache reuse grows with context length, developing methods that preserve accuracy for long inputs is essential.

In this paper, we demonstrate that combining model fine-tuning with selective recomputation of a small fraction of the KV caches enables more robust handling of long-context inputs.
We demonstrate that using both methods suppresses the deviation in the key entries (i.e., the values of the key vectors) more effectively than using either method alone because the two approaches contribute differently to reduce these deviations. 
As a result, our approach achieves a 9.7 point improvement in the RULER score over the CacheBlend-only approach at a 124k-token input.
Furthermore, TTFT is reduced by 80\% at a 124k-token input when the KV caches are loaded from the SSD.

\begin{figure}[t]
  \includegraphics[width=\columnwidth]{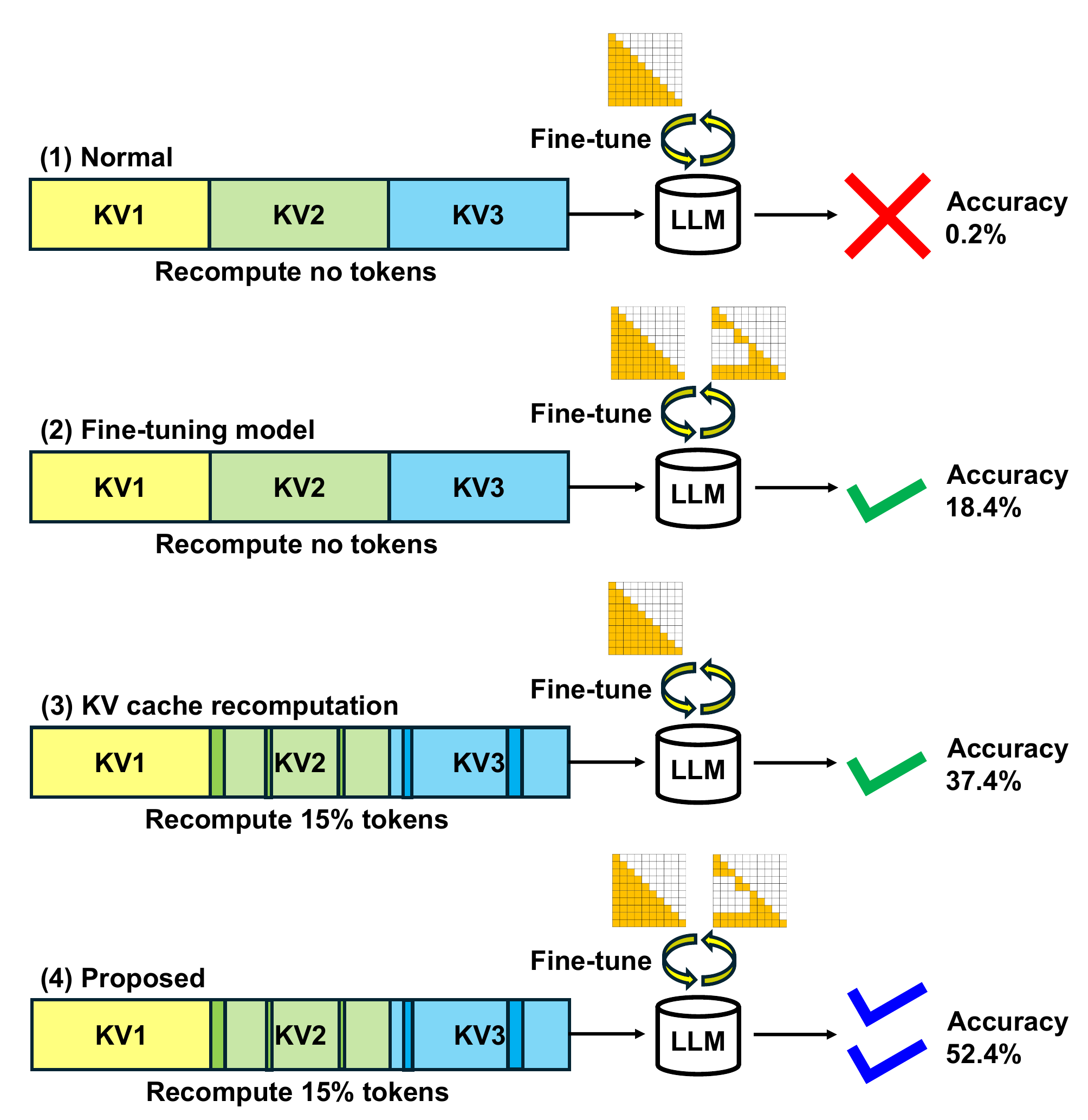}
  \caption{Comparison of four KV cache concatenation strategies: (1) normal, (2) fine-tuning model, (3) KV cache recomputation, and (4) the proposed combination.
           By combining fine-tuned model with selective KV cache recomputation, generation quality is better than using either technique alone.
           Accuracy values are taken from the results on the RULER HQA shown in Figure \ref{fig:ttft_accuracy_results_16packs}.}
  \label{fig:proposed_strategy}
\end{figure}

\section{Related Works}

\begin{figure}[t]
  \includegraphics[width=\columnwidth]{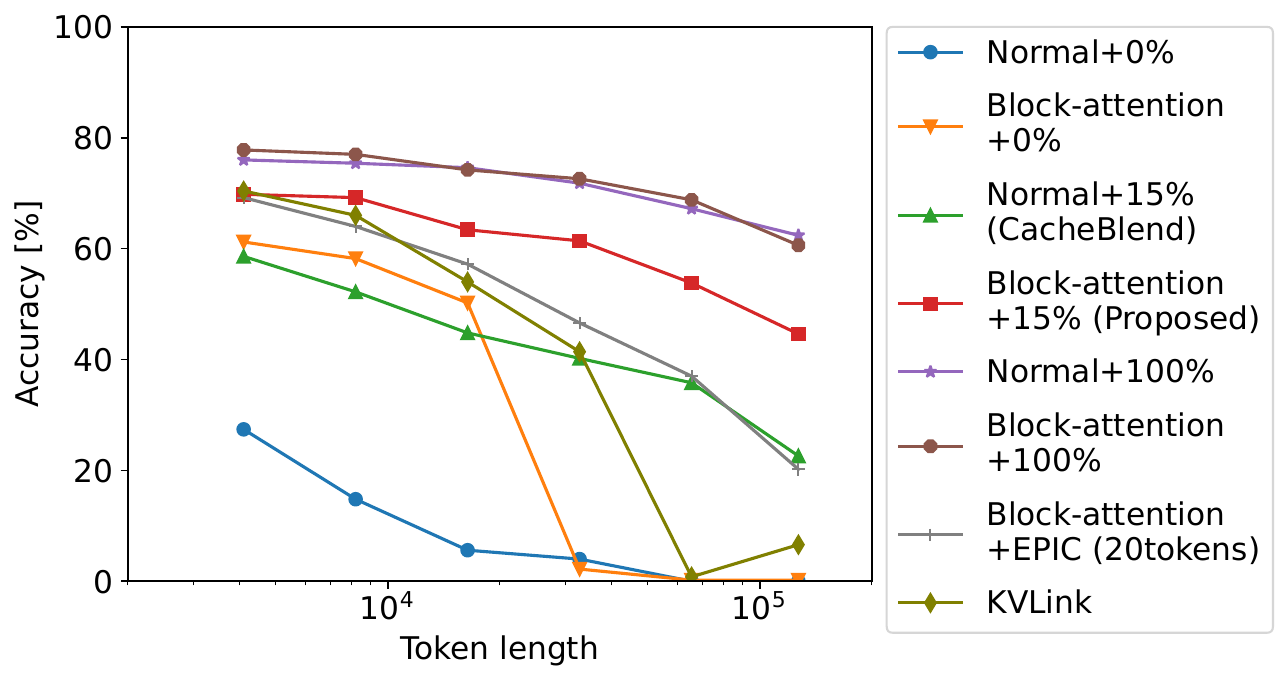}
  \caption{Evaluation results for the RULER HQA task with several approaches using Llama3.1-8B-Instruct. Although the fine-tuned-model only and recompute only approaches achieve comparable accuracy at a 4k-token input, the performance gap widens as the input context length increases.
           Among these approaches, the combination of Block-attention and CacheBlend (Proposed) achieves the best accuracy except for full attention.}
  \label{fig:ruler_hqa}
\end{figure}

A variety of studies have examined how to reuse and concatenate KV caches while maintaining generation quality.
Prompt Cache \cite{MLSYS2024_a66caa17} achieves this by applying a prompt markup language to merge KV caches.
PIE \cite{ICLR2025_95306350} addresses positional drift by aligning the concatenation point with the position expected by a model that uses a positional encoding scheme such as RoPE \cite{SU2024127063}.
Even with positional adjustment, quality degradation remains substantial because there is no cross-attention among reused KV caches from retrieved chunks.

Prior work addressing the quality degradation caused by the absence of cross-attention can be broadly grouped into three categories.
The first approach is to recompute KV caches for tokens selected according to a predefined rule.
CacheBlend \cite{yao2025cacheblend} recomputes the KV caches for token positions where the difference between the KV caches at the first layer obtained with full attention and the concatenated KV caches is large.
CacheClip \cite{yang2025cacheclipacceleratingrageffective} uses an auxiliary model to select token positions for recomputation.
EPIC \cite{hu2025epic} recomputes the KV caches for a fixed number of leading tokens in each chunk because attention scores tend to concentrate on the ''attention sink'' \cite{xiao2024efficient} portion of the chunk when the KV caches of the chunks are precomputed.
$\mathrm{A}^3$ \cite{zhou2025a3attentionawareaccuratekv} uses attention scores between each question token and the document tokens to select token positions for recomputation.
KVShare \cite{yang2025kvsharellmserviceefficient} uses attention scores and deviations in the value entries to select token positions for recomputation not only in the prefill phase but also in the decode phase.

\begin{figure*}[th]
  \centering
  \begin{subfigure}{0.42\linewidth}
    \centering
     \includegraphics[width=\linewidth]{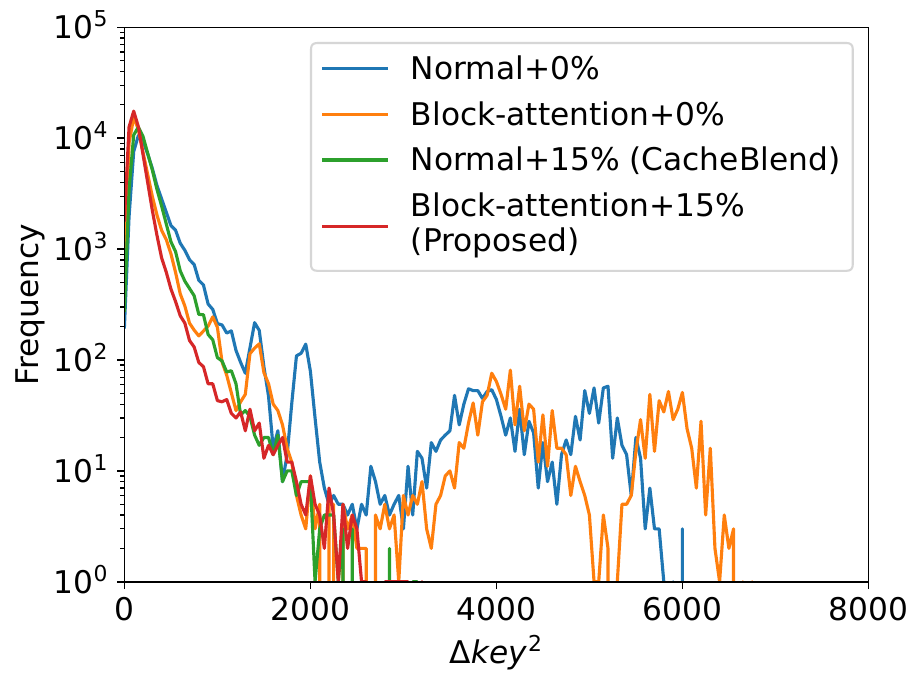}
     \caption{7th-layer}
  \end{subfigure}
  \begin{subfigure}{0.42\linewidth}
    \centering
     \includegraphics[width=\linewidth]{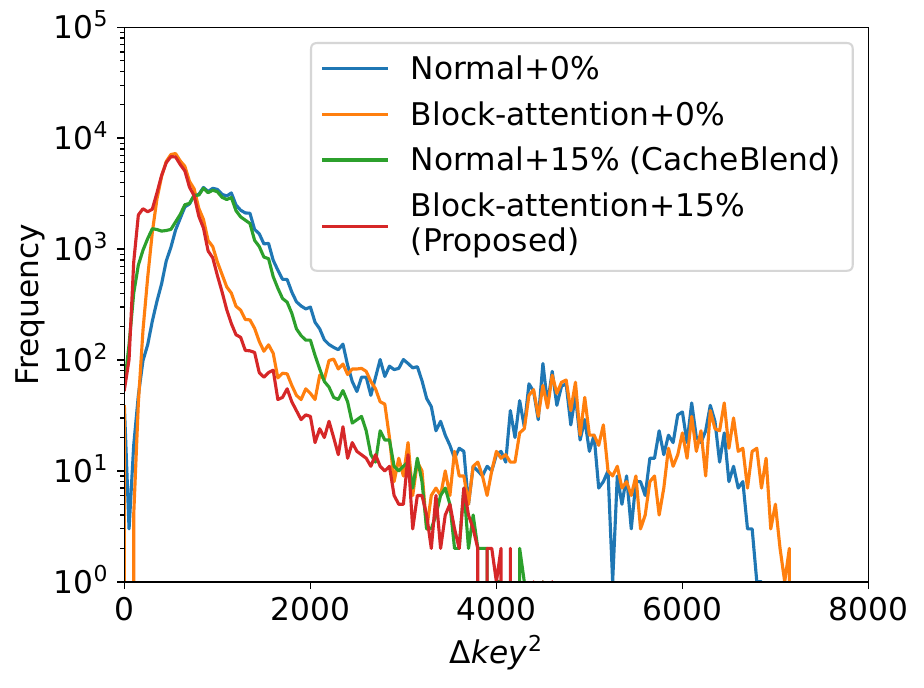}
     \caption{15th-layer}
  \end{subfigure} \\
  \begin{subfigure}{0.42\linewidth}
    \centering
     \includegraphics[width=\linewidth]{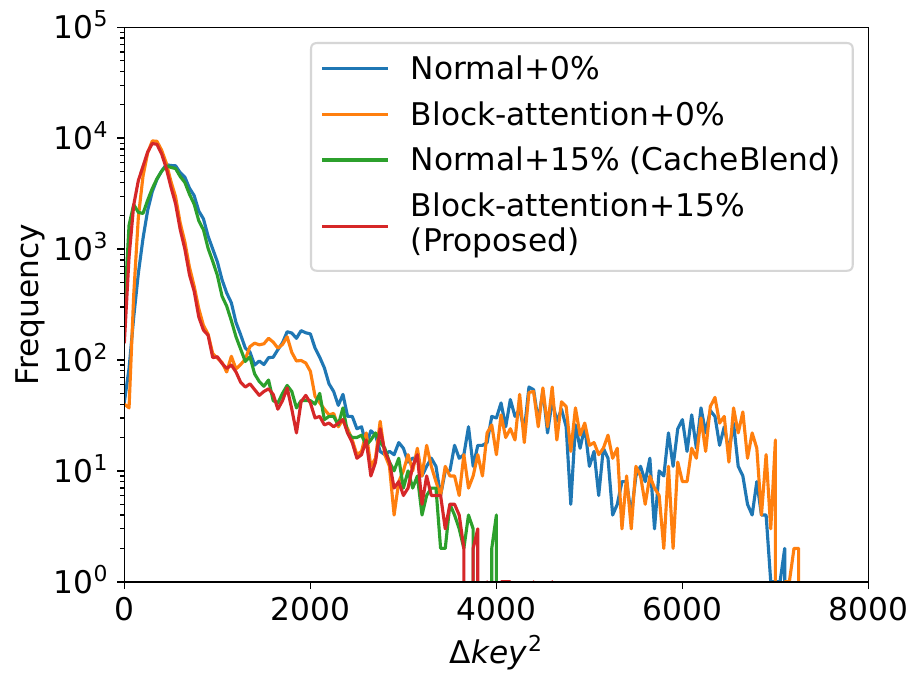}
     \caption{23rd-layer}
  \end{subfigure}
  \begin{subfigure}{0.42\linewidth}
    \centering
     \includegraphics[width=\linewidth]{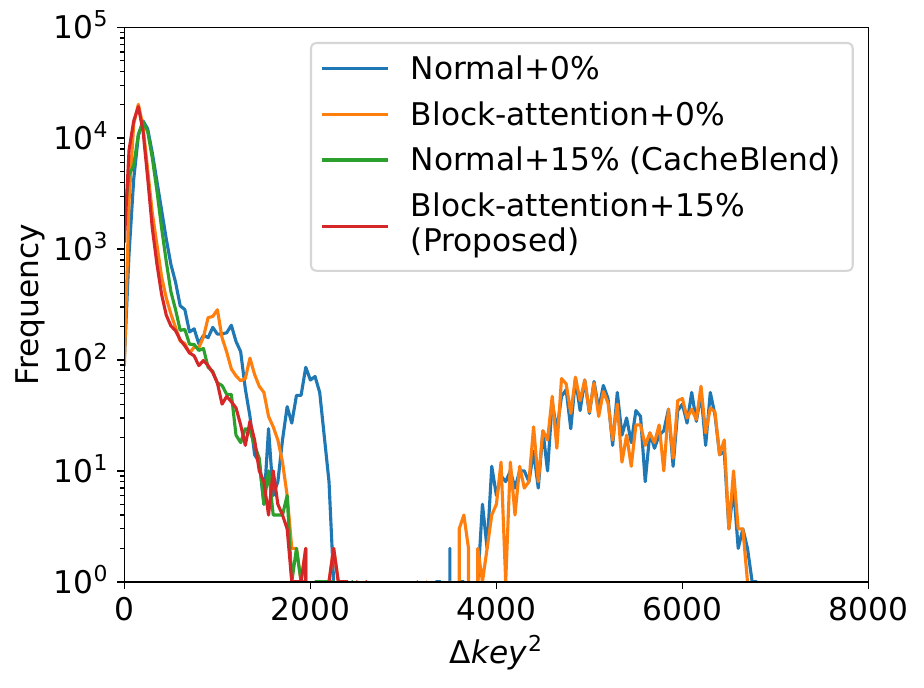}
     \caption{31st-layer}
  \end{subfigure} \\
  \caption{Squared differences in the key entries between those produced by the concatenated cache approach and those obtained with full attention ($\Delta key^2$) at layers 7 through 31 using Llama3.1-8B-Instruct (64k-token input).
           While the Block-attention-only approach produces a steeper distribution with a peak shifted further to the left than that of the Normal approach, the CacheBlend-only approach reduces large deviations mainly induced by special tokens for chunk separation and the attention sink that occurs within chunks.
           The combination of both approaches (Proposed) inherits both characteristics.}
  \label{fig:kdiff_llama}
\end{figure*}

The second approach is to fine-tune the model under the condition that cross-attention is absent between chunks so that the LLM can preserve generation quality.
Block-attention \cite{ma2025blockattention} and TurboRAG \cite{lu-etal-2025-turborag} fine-tune the model so that concatenated KV caches can be reused without significant degradation.
KVLink \cite{yang2025kvlink} inserts trainable special tokens between chunks during model fine-tuning. These tokens have cross-attention to capture information from the preceding tokens.
Apart from the fine-tuning step, KVLink is similar to EPIC.

The third approach focuses on differences in attention scores between KV caches computed with full attention and concatenated KV caches.
Link0 \cite{hu2025epic} places prefix tokens before each chunk when precomputing KV caches to mitigate the effect of the attention sink.
In addition to placing prefix tokens, APE \cite{ICLR2025_ce186a37} computes the softmax for the context with a lower temperature and multiplies the attention scores by an another scaling factor.

\citet{cestola2026experimentalstudykvcache} reported an experimental evaluation of existing approaches.
Then they proposed combining CacheBlend, Link0, and APE to achieve better accuracy. However, the combination of a fine-tuned model and other approaches has not been examined.

Although the aforementioned techniques have been evaluated on several tasks, few works have examined how response accuracy depends on input context length.
Reusing KV caches for long-context inputs can significantly benefit TTFT because the prefill recomputation cost scales as $O(n^2)$. 
Existing works tend to report speed gains for long contexts but provide little empirical evidence regarding accuracy.

\section{Methodology}

\subsection{Issues in existing approaches: response quality at long context inputs}
Figure \ref{fig:ruler_hqa} shows results on a synthetic HotpotQA (HQA) \cite{yang-etal-2018-hotpotqa} task generated by RULER \cite{hsieh2024ruler}, comparing Block-attention, CacheBlend with a 15\% recomputation rate, EPIC, and KVLink.
Two-token special tokens are inserted between chunks, and the KV caches at these positions are set to zero.
KVLink recomputes the special tokens at the end of each chunk, whereas EPIC recomputes the 20 tokens starting from that position.
To match the evaluation setup of the HQA task executed with Block-attention, we use the same prompt as Block-attention, and we use the $\mathrm{best\_subspan\_em}$ metric for evaluation following \citet{liu-etal-2024-lost}.
Normal denotes a model fine-tuned with the standard approach using the same dataset as Block-attention and KVLink for comparison.
The input lengths are 4k, 8k, 16k, 32k, 64k, and 124k tokens, respectively.
The 124k setting corresponds to 126,976 tokens, reduced from the original 128k setting (131,072 tokens) so that the final input context length after adding special tokens does not exceed the max token length of the model.
At a 4k-token input, the accuracy is close to that of full attention.
However, as the context length increases, the accuracy gap widens, indicating that current methods do not guarantee sufficient accuracy for very long inputs, where reusing KV caches is an effective way to reduce TTFT.

\subsection{Motivation of proposed method: effect of fine-tuning/recomputing on the deviations in the key entries}
Fine-tuning approaches such as Block-attention make the model more robust to KV cache concatenation by incorporating the assumption that there is no cross-attention between documents into the fine-tuning training data.
On the other hand, recomputing methods such as CacheBlend select a set of token positions according to a predetermined rule and recompute only those KV caches.
We investigate the effects of the two approaches in terms of deviations in the key entries compared with those obtained with full attention.
Figure \ref{fig:kdiff_llama} shows the distribution of squared differences in the key entries between those produced by the concatenated cache approach and those obtained with full attention across three experimental conditions: (1) Normal, (2) Block-attention-only, and (3) CacheBlend-only.
These distributions are obtained from the first task of the RULER HQA with 64k-token input.
In our experiments, CacheBlend selects token indices based on the squared differences in the key entries, as implemented in LMCache\footnote{https://github.com/LMCache/LMCache} \cite{liu2025lmcacheefficientkvcache}.
Note that two-token special tokens are inserted between chunks, and the KV caches at these special token positions are set to zero to follow the CacheBlend implementation in vLLM \cite{kwon2023pagedattention} + LMCache; the recomputation rate is 15\%.
As shown in Fig. \ref{fig:kdiff_llama}, the Block-attention-only approach produces a steeper distribution with a peak shifted further to the left than the Normal approach.
On the other hand, the CacheBlend-only approach reduces large deviations.
The differences in the distributions arise from the fact that CacheBlend-only checks large deviations in the key entries at first-layer, while Block-attention fine-tunes the model parameters across all layers.  
Combining both approaches can benefit from the strengths of each method because the two approaches contribute differently to reduce deviations in the key entries.
We therefore also investigated the additional condition: (4) Block-attention+CacheBlend.
As shown in Fig. \ref{fig:kdiff_llama}, the Block-attention+CacheBlend approach inherits the strengths of both approaches and achieves the smallest deviations among them.
These results suggest that combining Block-attention and CacheBlend can achieve higher accuracy than either method alone, as shown in Figure \ref{fig:proposed_strategy}.

\subsection{Other Candidates for KV Cache Recomputation and Performance Comparison}
\begin{figure}[t]
  \includegraphics[width=\columnwidth]{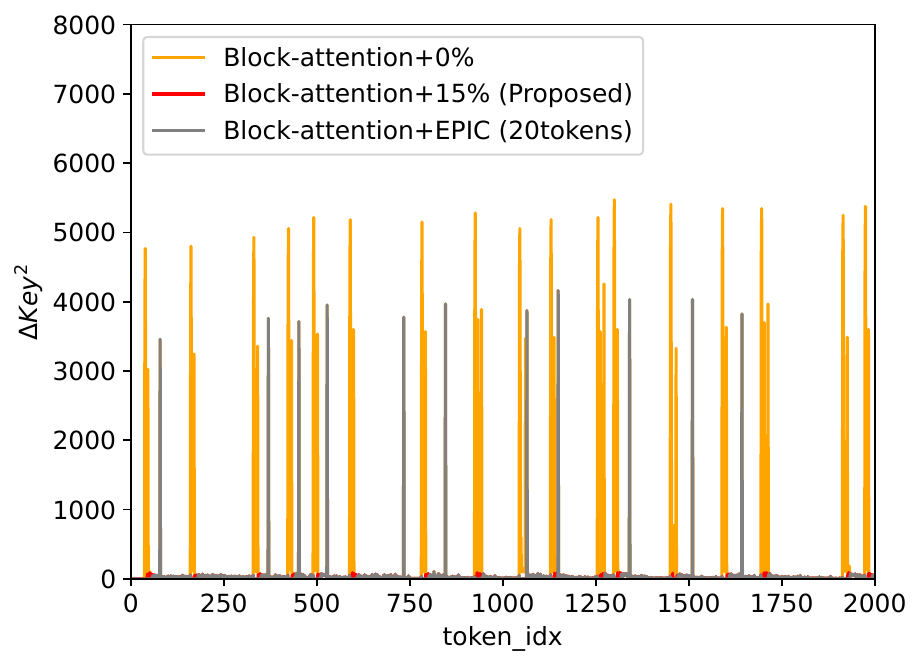}
  \caption{The distribution of squared differences in the key entries with no recomputation (0\%), CacheBlend (15\%), and EPIC at the first layer using Llama3.1-8B-Instruct. Because EPIC recomputes KV caches only around the attention sink of each chunk, it cannot reduce large deviations far from the attention sink.}
  \label{fig:epic_vs_cacheblend}
\end{figure}

The combination of Block-attention and EPIC is also a candidate for improving performance, but the improvement may be smaller than that achieved with CacheBlend.
Figure \ref{fig:epic_vs_cacheblend} shows key-value cache differences at the first layer using EPIC and CacheBlend.
Although CacheBlend recomputes the KV caches for token positions with large deviations in the key entries, EPIC recomputes the KV caches around the attention sink of each chunk.
As a result, large deviations in the key entries far from the attention sink remain unrecomputed.
These deviations could cause performance degradation.

Using attention scores obtained from the query, as in $\mathrm{A}^3$ and KVShare, is also an attractive candidate for improving performance.
However, our experiments using KVShare did not show any measurable improvement under long input as shown in Appendix \ref{sec:appendix_kvshare}, so we did not include these methods from the main evaluation.
Moreover, Link0 and APE could potentially improve performance, but we did not test their combination in this paper.

As shown in Fig. \ref{fig:ruler_hqa}, the combination of Block-attention and CacheBlend achieves the best accuracy among the evaluated strategies.
Therefore, we selected this combination for our evaluation.
A fine-tuned model can be combined with CacheBlend's KV cache recomputation in the same way as a non-fine-tuned model.
As a result, the inference latency remains essentially the same as that of CacheBlend alone, while the accuracy improves.

\subsection{Implementation}
\begin{figure}[t]
  \includegraphics[width=\columnwidth]{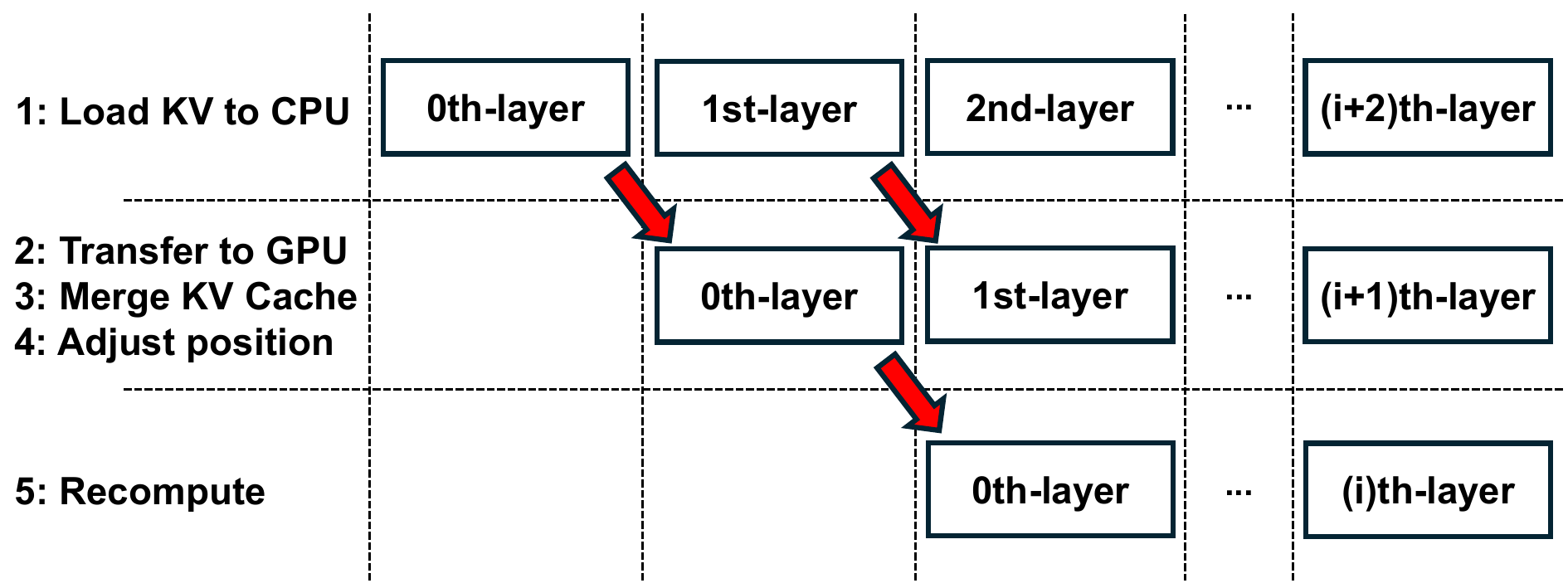}
  \caption{Timing chart of layer-wise KV cache loading and recomputation. Loading KV caches to the CPU, transferring KV caches to the GPU and merging with position adjustment, and recomputation are executed in parallel.}
  \label{fig:load_and_save}
\end{figure}

We implemented a layer-wise parallel operation for loading KV caches and recomputing them, as shown in Figure \ref{fig:load_and_save}.
This operation consists of three parts.
First, precomputed KV caches for the selected layer are loaded from safetensors-format files on the SSD (num\_worker=8).
Second, the loaded KV caches are transferred to the GPU, and the KV caches for the chunks are merged into one, followed by position adjustment of the KV caches.
Finally, the transformer-layer computation, including KV cache recomputation, is performed.
The first part runs with ThreadPoolExecutor, and the second and third parts use CUDA streams to overlap computation and KV cache transfer.

\section{Experiments}
\label{experiments}
This section evaluates the performance of the proposed approach.
We used transformers v4.57.1 for both training and inference.
First, we use Llama3.1-8B-Instruct \cite{grattafiori2024llama3herdmodels} as the base model and fine-tune it to be aware of KV cache concatenation, following the training datasets for Block-attention in \cite{ma2025blockattention}.
The dataset used for fine-tuning consists of two parts.
The first part consists of 40,000 samples from Tulu3 \cite{lambert2025tulu}.
The second part is created by selecting 21,000 samples each from TriviaQA (TQA) \cite{joshi-etal-2017-triviaqa} and 2WikiMultiHopQA (2Wiki) \cite{ho-etal-2020-constructing}.
Then, ten documents are retrieved randomly for each sample using the Contriever toolkit\footnote{https://github.com/facebookresearch/contriever}, and shuffling the order of samples.
After that, the generated answers for these samples are produced with GPT-4o mini. Finally, samples where GPT-4o mini fails to generate an answer are excluded, and 40,000 samples are selected for the training dataset.
The contexts in these datasets are split into chunks in the same way as Block-attention\footnote{https://github.com/temporarylora/block-attention} to obtain the Block-attention model. 
Fine-tuning is performed on a dual-socket system with four NVIDIA H200 141GB GPUs and two Intel(R) Xeon(R) Platinum 8562Y+ CPUs, and inference is performed on a single GPU.
The model is fine-tuned on the concatenation of these two datasets using a learning rate of $2 \times 10^{-6}$, a batch size of four (one per GPU), gradient accumulation steps of 16, one epoch, a warmup ratio of 0.03, and the AdamW optimizer.
DeepSpeed\footnote{https://github.com/deepspeedai/deepspeed} is used to accelerate the training procedure using the bfloat16 format.
FlashAttention-2 \cite{dao2024flashattention} is also used for the Normal.
While the Normal model is fine-tuned using the dataset only with full attention, Block-attention uses the dataset both with full attention and under the assumption that there is no cross-attention between chunks.
It takes about five hours to create the Block-attention model from Llama3.1-8B-Instruct.
In our experiments, KVLink uses special tokens for chunk separation as trainable special tokens during fine-tuning.

During inference, we perform the recomputation stage using FlashInfer \cite{ye2025flashinferefficientcustomizableattention} with a custom attention mask.
Position adjustment for KV cache concatenation is performed in float32 while other computations in the model are performed using the bfloat16 format.
We perform text generation with a single run greedy decoding (do\_sample=False, repetition\_penalty=1.0), setting max\_new\_tokens to 512 for all tasks except the RULER benchmark, where each task has its own specified value.
Note that full recomputation, such as for the 0th layer, is automatically switched to causal attention mode.
The prefill stage for instruction tokens and the decode stage are processed with FlashAttention-2.
To facilitate future deployment with LMCache, we inserted two-token special tokens between documents and filled the KV caches corresponding to those special token indices with all-zero values.
From the original RULER benchmark \cite{hsieh2024ruler}, all passages except those for QA tasks were concatenated into documents of no more than 512 tokens and adapted for Block-attention and CacheBlend.
For QA tasks, we use the pre-existing documents directly as individual chunks.

\subsection{Generation Quality on the RULER Benchmark}

\begin{table*}
\centering
\small
\begin{tabular}{lcccccccccc}
\hline
Tasks & Single & MK & MV & MQ & VT & CWE & FWE & QA1 & QA2 & Avg. \\
\hline
Normal+0\% & 98.7           & 79.6          & 43.0          & 74.5          & 31.8           & 86.7           & 95.3          & 36.0          & 28.0          & 71.6          \\
Normal+15\% (CacheBlend) & \textbf{100.0} & 92.9          & \textbf{56.8} & 84.5          & 53.1           & \textbf{97.8}  & \textbf{98.5} & 74.0          & 48.8          & 84.0          \\
Normal+100\% & 100.0          & 99.9          & 100.0         & 100.0         & 100.0          & 98.6           & 88.8          & 82.4          & 63.4          & 94.8          \\
\hline
Block-attention+0\% & 99.9           & 93.7          & 53.1          & \textbf{89.8} & 51.8           & 86.8           & 95.3          & 69.6          & 47.4          & 82.7          \\
Block-attention+15\% (Proposed) & \textbf{100.0} & \textbf{96.9} & 56.1          & 86.7          & \textbf{70.8}  & 96.7           & 95.3          & \textbf{75.4} & \textbf{52.8} & \textbf{86.5} \\
Block-attention+100\% & 100.0          & 99.9          & 89.5          & 100.0         & 100.0          & 98.9           & 85.3          & 81.4          & 61.4          & 93.5          \\
\hline
\end{tabular}
\caption{Evaluation results for the RULER benchmark at 4k-token input using Llama3.1-8B-Instruct. Single and MK denote the average score of Single NIAH 1-3, and Multi-Key NIAH 1-3, respectively.
         Avg. denotes the average score across all 13 tasks. At a 4k-token input, the Block-attention-only and CacheBlend-only approaches achieve scores comparable to those of the proposed approach.}
\label{tab:ruler_4k_results}
\end{table*}

\begin{table*}
\centering
\small
\begin{tabular}{lcccccccccc}
\hline
Tasks & Single & MK & MV & MQ & VT & CWE & FWE & QA1 & QA2 & Avg. \\
\hline
Normal+0\% & 0.0 & 0.0 & 0.0 & 0.0 & 0.0 & 0.0 & 22.3 & 1.6 & 1.0 & 1.9 \\
Normal+15\% (CacheBlend) & 93.3 & 22.1 & \textbf{45.8} & 61.5 & 35.0 & 0.0 & 72.7 & 29.4 & 20.0 & 47.0 \\
Normal+100\% & 99.9 & 90.3 & 95.6 & 98.0 & 74.5 & 0.1 & 79.5 & 77.2 & 47.4 & 80.2 \\
\hline
Block-attention+0\% & 3.2 & 2.1 & 0.7 & 0.6 & 0.0 & 0.0 & 51.9 & 2.4 & 8.8 & 6.2 \\
Block-attention+15\% (Proposed) & \textbf{98.3} & \textbf{35.8} & 42.0 & \textbf{75.6} & \textbf{57.1} & \textbf{0.1} & \textbf{82.1} & \textbf{48.4} & \textbf{29.4} & \textbf{56.7} \\
Block-attention+100\% & 100.0 & 92.1 & 94.9 & 98.2 & 80.9 & 0.1 & 81.7 & 75.0 & 47.8 & 81.1 \\
\hline
\end{tabular}
\caption{Evaluation results for the RULER benchmark at a 124k-token input using Llama3.1-8B-Instruct. At a 124k-token input, the proposed approach achieves the best score on most tasks compared with approaches that use either technique alone.}
\label{tab:ruler_124k_results}
\end{table*}

\begin{figure}[t]
  \includegraphics[width=\columnwidth]{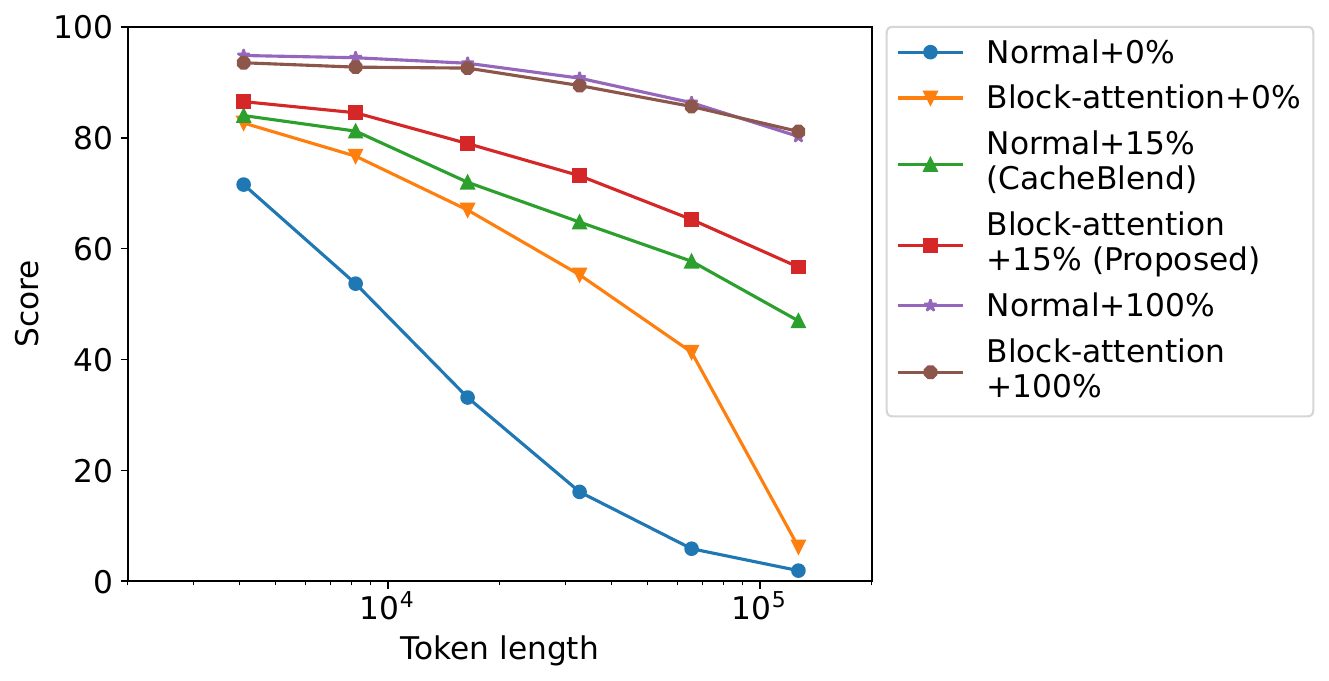}
  \caption{Evaluation results for the RULER benchmark across several input context lengths using Llama3.1-8B-Instruct. The proposed approach shows the smallest degradation from full attention (100\%) compared with approaches that use either technique alone.}
  \label{fig:ruler_after_proposal}
\end{figure}

\begin{figure}[t]
  \includegraphics[width=\columnwidth]{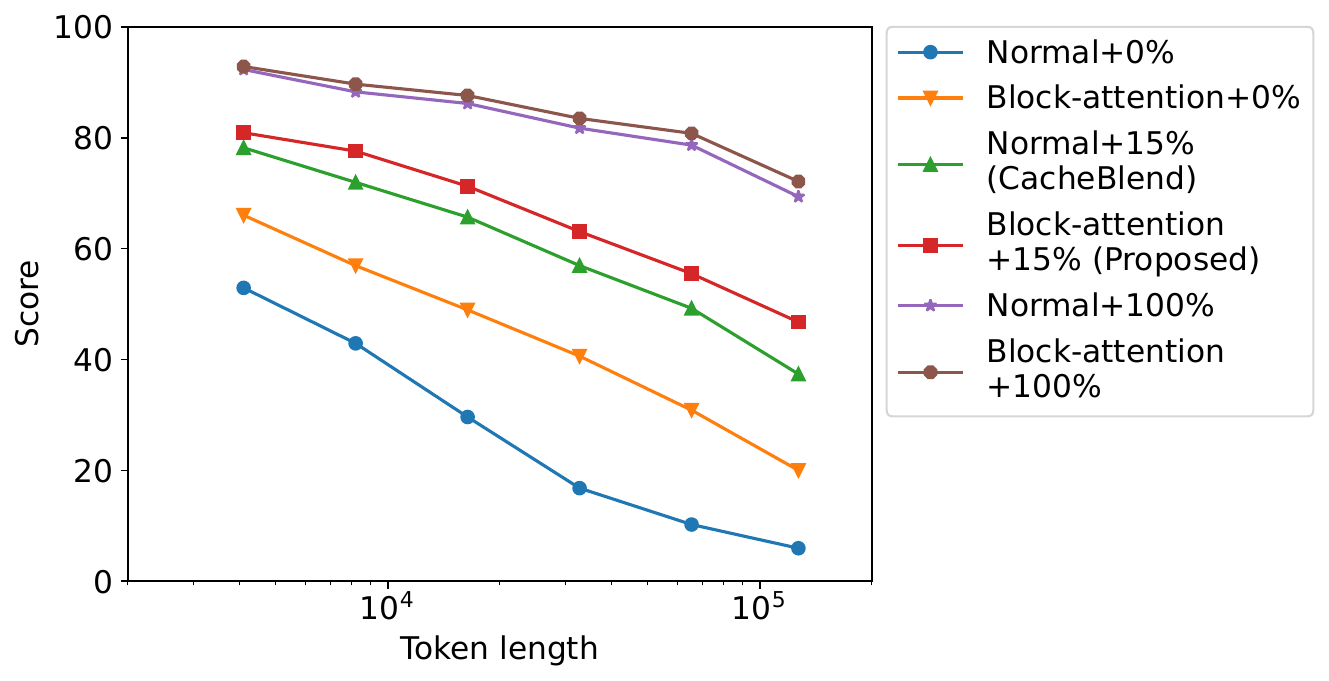}
  \caption{Evaluation results for the RULER benchmark across several input context lengths using Qwen2.5-7B-Instruct. Results consistent with those for Llama3.1-8B-Instruct are obtained.}
  \label{fig:ruler_after_proposal_qwen2}
\end{figure}

\begin{figure}[t]
  \includegraphics[width=\columnwidth]{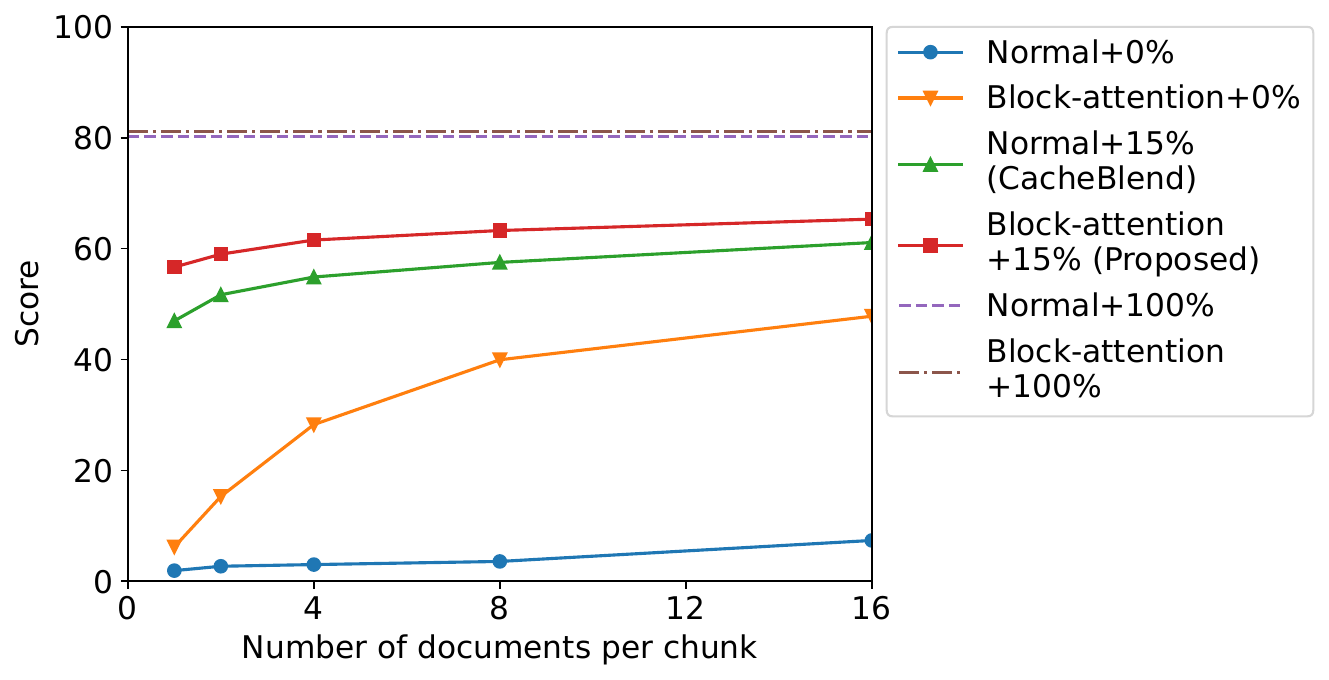}
  \caption{Evaluation results for the RULER benchmark with different numbers of documents per chunk at a 124k-token input using Llama3.1-8B-Instruct. As the number of chunks decreases, the RULER score recovers toward the full attention (100\%) result. Note that the full attention result is obtained with one document per chunk.}
  \label{fig:ruler_changing_document_chunks}
\end{figure}

Figure \ref{fig:ruler_after_proposal} shows the evaluation results for the RULER benchmark using Llama3.1-8B-Instruct.
Tables \ref{tab:ruler_4k_results} and \ref{tab:ruler_124k_results} show the RULER scores of several tasks at 4k and 124k tokens.
This figure demonstrates that as the number of input context tokens grows, the score gap relative to the 100\% recomputation baseline (equivalent to full attention) widens.
Among the evaluated variants, the Block-attention configuration with a 15\% recomputation rate shows the smallest accuracy degradation, indicating that combining fine-tuning with a selective recomputation strategy improves scores in scenarios that merge KV caches.

Figure \ref{fig:ruler_after_proposal_qwen2} reports the same set of experiments performed with Qwen2.5-7B-Instruct model \cite{qwen2025qwen25technicalreport}.
This model uses YARN \cite{peng2024yarn}, and the learning rate is $5 \times 10^{-6}$. It takes about 7.5 hours to create the Block-attention model from Qwen2.5-7B-Instruct.
The results also confirm that Block-attention combined with CacheBlend delivers the best performance with a different model.

Note that the number of chunks increases as the number of tokens grows in the experiments shown in Figs. \ref{fig:ruler_after_proposal} and \ref{fig:ruler_after_proposal_qwen2}.
As a result, the number of chunks at a 124k-token input exceeds 200 (and 700 in the case of QA tasks).
To evaluate the effect of the number of chunks alone, we also evaluate the RULER benchmark under the condition in which the number of documents per chunk is changed, reducing the number of chunks while keeping the number of tokens unchanged except for the special tokens between chunks.
Figure \ref{fig:ruler_changing_document_chunks} shows the impact of the number of documents per chunk on the RULER score at a 124k-token input.
When the number of documents per chunk increases from 1 to 16, the score gap between full attention and the proposed approach decreases from 24.5 to 15.8.
This means that increasing the number of documents per chunk can improve the RULER score, but at the cost of loading redundant chunks or lowering the retrieval hit rate because the minimum token length per chunk increases.
On the other hand, the RULER score improvement of the proposed approach from the CacheBlend-only approach is larger as the number of documents per chunk is smaller.
This result shows that the proposed approach is effective at maintaining generation quality even with a higher number of chunks.

\subsection{Performance in RAG and General Tasks}
Table \ref{tab:rag_results} shows the evaluation results for four RAG tasks. We use the same prompt as in Fig. \ref{fig:ruler_hqa}.
The four benchmarks are 2Wiki \cite{ho-etal-2020-constructing}, HQA \cite{yang-etal-2018-hotpotqa}, Natural Questions (NQ) \cite{kwiatkowski-etal-2019-natural}, and TriviaQA (TQA) \cite{joshi-etal-2017-triviaqa}.
Ten documents are retrieved for each task using the Contriever toolkit.
$\mathrm{best\_subspan\_em}$ is used as the evaluation metric, just as in Fig. \ref{fig:ruler_hqa}.
The proposed approach achieves the best accuracy among them, except for full attention.
Table \ref{tab:opencompass_results} shows the evaluation results for general-purpose tasks using OpenCompass \cite{2023opencompass}.
As in the experiments shown for Block-attention, in-context learning (ICL) datasets such as GSM8K \cite{cobbe2021trainingverifierssolvemath}, MATH \cite{hendrycks2021measuring_b}, BBH \cite{suzgun2022challengingbigbenchtaskschainofthought}, and DROP \cite{dua-etal-2019-drop} are evaluated with KV cache concatenation.
Zero-shot datasets such as MMLU \cite{hendrycks2021measuring_a}, HumanEval \cite{chen2021evaluatinglargelanguagemodels}, and IFEval \cite{zhou2023instructionfollowingevaluationlargelanguage} are evaluated with full attention.
For IFEval, prompt-level-strict-accuracy is selected for evaluation.
As shown in Table \ref{tab:opencompass_results}, the proposed approach lies between the 0\% and 100\% recomputation settings, and selective recomputation is beneficial even for general tasks.

\begin{table*}
\centering
\small
\begin{tabular}{lcccc}
\hline
Approaches & 2wiki & HQA & NQ & TQA \\
\hline
\hline
Normal+0\% & 42.0          & 42.2          & 41.9          & 63.1          \\
Normal+15\% (CacheBlend) & 65.9          & 66.1          & 57.1          & 73.8          \\
Normal+100\% & 79.5          & 78.0          & 62.6          & 77.3          \\
\hline
Block-attention+0\% & 68.8          & 70.1          & 57.1          & 74.5          \\
Block-attention+15\% (Proposed) & \textbf{70.8} & \textbf{72.6} & \textbf{58.1} & \textbf{75.2} \\
Block-attention+100\% & 78.5          & 77.8          & 61.7          & 77.0          \\
\hline
\end{tabular}
\caption{Evaluation results on four RAG benchmarks using Llama3.1-8B-Instruct. Among the approaches other than full attention, the proposed approach achieves the best accuracy.}
\label{tab:rag_results}
\end{table*}

\begin{table*}
\centering
\small
\begin{tabular}{lccccccc}
\hline
Task Type & \multicolumn{3}{c}{General} \vline & \multicolumn{4}{c}{ICL} \\
\hline
dataset & \multicolumn{1}{l}{IFEval} & \multicolumn{1}{l}{HumanEval} & \multicolumn{1}{l}{MMLU} & \multicolumn{1}{l}{GSM8K} & \multicolumn{1}{l}{MATH} & \multicolumn{1}{l}{BBH} & \multicolumn{1}{l}{DROP} \\
setup & \multicolumn{1}{l}{0-shot} & \multicolumn{1}{l}{0-shot} & \multicolumn{1}{l}{0-shot} & \multicolumn{1}{l}{4-shot} & \multicolumn{1}{l}{4-shot} & \multicolumn{1}{l}{3-shot} & \multicolumn{1}{l}{3shot} \\
\hline
\hline
Normal+0\% &                &               &               & 65.1          & 24.3          & 61.9          & 12.4          \\
Normal+15\% (CacheBlend) & 64.3          & 57.3          & \textbf{66.0} & 76.3          & \textbf{35.9} & \textbf{70.6} & \textbf{13.3} \\
Normal+100\% &              &               &               & 79.3          & 35.0          & 70.8          & 15.3          \\
\hline
Block-attention+0\% &                &               &               & 75.1          & 32.4          & 68.1          & 12.3          \\
Block-attention+15\% (Proposed) & \textbf{66.0} & \textbf{64.0} & 65.9          & \textbf{78.2} & 35.7          & \textbf{70.6} & 12.3          \\
Block-attention+100\% &              &               &               & 79.5          & 34.9          & 70.4          & 13.2          \\
\hline
\end{tabular}
\caption{Evaluation results for seven general benchmarks using Llama3.1-8B-Instruct. Only the four ICL tasks with few-shot examples are split into different chunks. Compared with no approach, all three methods improve performance on the ICL tasks.}
\label{tab:opencompass_results}
\end{table*}

\subsection{Token-Length Dependence of TTFT with the Proposed Approach}

\begin{figure}[t]
  \includegraphics[width=\columnwidth]{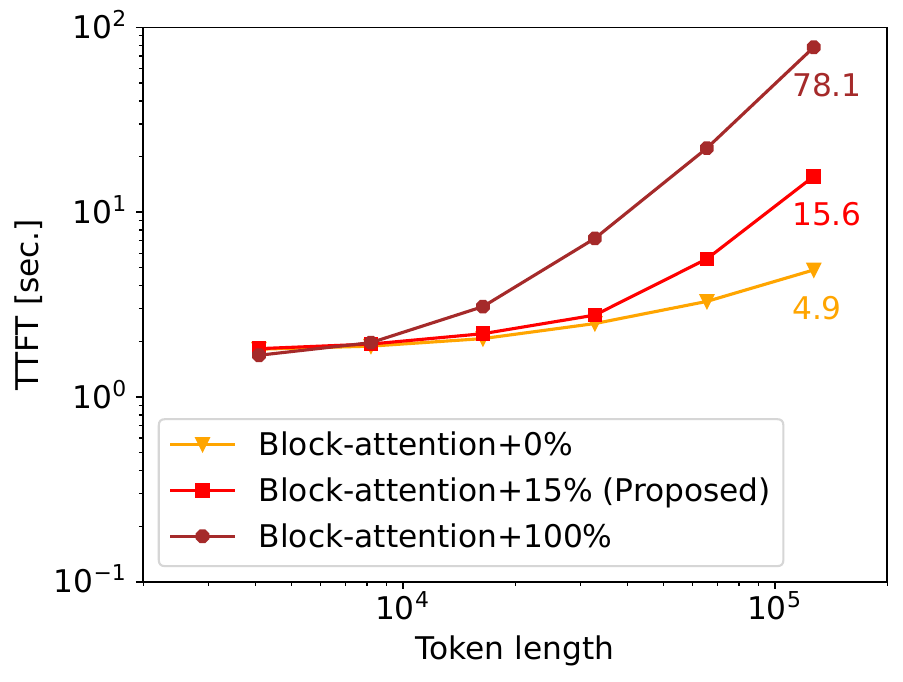}
  \caption{Averaged TTFT over 20 tasks in the RULER HQA using Llama3.1-8B-Instruct with the number of documents per cache file set to 16.
           The Proposed approach achieves much lower TTFT than full attention (100\%).}
  \label{fig:token_ttft_results_16packs}
\end{figure}

\begin{figure}[t]
  \includegraphics[width=\columnwidth]{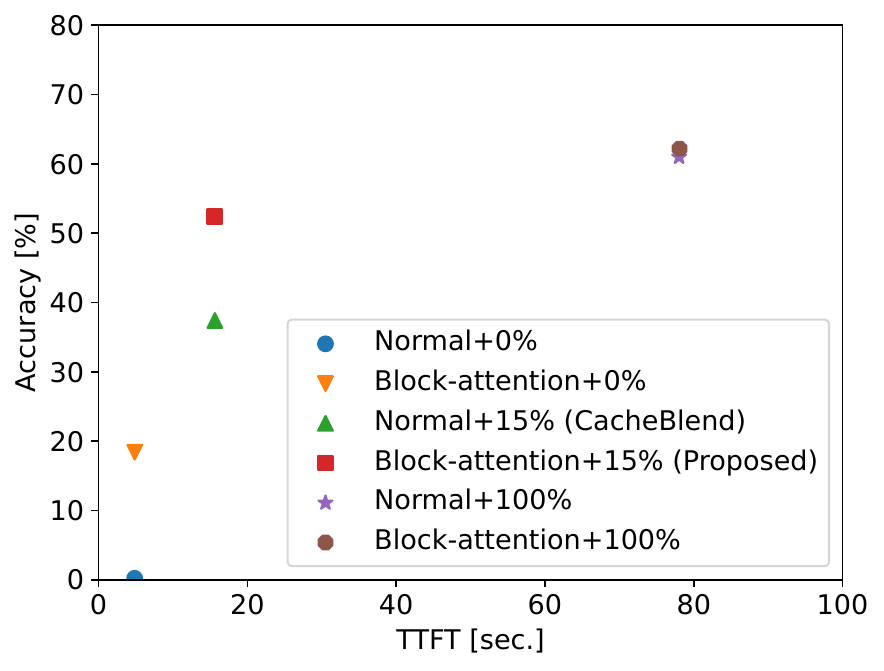}
  \caption{Relationship between averaged TTFT over 20 tasks and the accuracy in the RULER HQA at a 124k-token input using Llama3.1-8B-Instruct with the number of documents per cache file set to 16.}
  \label{fig:ttft_accuracy_results_16packs}
\end{figure}

These results demonstrate that the combination of Block-attention and CacheBlend can substantially improve performance while reducing computational overhead.
Figure \ref{fig:token_ttft_results_16packs} plots the relationship between input context length and TTFT.
All plots include the running time for approximately 1.5 s to execute \texttt{model.generate} after KV cache preparation.
In this experiment, FlashInfer with an attention mask is used for every attention computation to measure the effect of the recomputation rate.
The KV caches are pre-stored on an SSD with a bandwidth of 6.5 GB/s, measured using \texttt{fio}.
For QA tasks, the cache files corresponding to the task are loaded from the SSD.
When the token length per file is short, the overhead incurred per file becomes a large fraction of the actual data read time, thereby diminishing the benefit of reusing KV cache.
Accordingly, we set the number of documents per cache file to 16, resulting in an average token count of about 2.2k per file.
For a context length of 4k tokens, full prefill computation yields a shorter TTFT than loading KV caches.
However, once the context length reaches 8k tokens or more, loading KV caches becomes advantageous over full prefill computation, and TTFT is reduced by 80\% at a 124k-token input.
Figure \ref{fig:ttft_accuracy_results_16packs} shows the relationship between TTFT and the RULER HQA accuracy under this condition.
The use of a Block-attention fine-tuned model achieves an accuracy gain over the Normal model without additional TTFT cost.

\section{Conclusions}
We addressed the degradation in generation quality that occurs when precomputed KV caches are concatenated for long-context inputs, especially when KV cache loading becomes more advantageous than prefill computation in terms of TTFT.
To reduce this performance degradation, we proposed combining KV cache-concatenation-aware fine-tuning with recomputation of selected KV caches.
Experimental results showed that the proposed technique reduces degradation in generation quality for long-context inputs compared with either existing technique alone.
We also demonstrated that TTFT is reduced by 80\% at a 124k-token input when the KV caches are loaded from the SSD.

\section*{Limitations}
This section discusses the limitations of this paper that we intentionally leave as future work.

\textbf{Limitation 1: Computation time of attention that supports recomputation for selected query indices.}
In our experiments, FlashInfer with an attention mask is used to compare selective recomputation with full attention under the same attention computation setting.
However, causal attention can use FlashAttention-2 to further reduce computation time.
If only selective recomputation uses FlashInfer with an attention mask while full attention uses FlashAttention-2, the benefit of selective recomputation plus KV cache loading decreases as shown in Appendix \ref{sec:appendix_ttft}.
However, selective recomputation without an attention mask incurs attention to future KV positions, which results in quality degradation, especially for long-context inputs.
To fully exploit selective recomputation, a faster attention mechanism that supports selected query indices is required.

\textbf{Limitation 2: Evaluation of attention-score-based recomputation.}
In our experiments, we did not evaluate attention-score-based recomputation methods such as $\mathrm{A}^3$ and KVShare.
However, the use of these recomputation methods should also be evaluated because these methods could achieve better performance than CacheBlend.
While the evaluation results for the RULER benchmark using KVShare as shown in Appendix \ref{sec:appendix_kvshare}, detailed evaluation is required.

\textbf{Limitation 3: Evaluation of an additional combination of Link0 and APE.}
We did not evaluate the combination of Link0 and APE because we focused on the combination of KV cache recomputation and model fine-tuning.
However, as stated in \citeauthor{cestola2026experimentalstudykvcache}\citeyearpar{cestola2026experimentalstudykvcache}, an additional combination of Link0 and APE could help achieve even better performance.

\textbf{Limitation 4: Fine-tuning thinking models.}
In this paper, we did not fine-tune thinking models such as Qwen3 \cite{yang2025qwen3technicalreport} and gpt-oss \cite{openai2025gptoss120bgptoss20bmodel}.
A new fine-tuning strategy that incorporates a Block-attention-style design without reducing chain-of-thought capability is required.


\bibliography{custom}

\appendix

\section{Detailed Results for NIAH tasks}
\label{sec:appendix_niah}

Tables \ref{tab:ruler_single_niah_4k_results}, \ref{tab:ruler_multikey_niah_4k_results}, \ref{tab:ruler_single_niah_124k_results}, and \ref{tab:ruler_multikey_niah_124k_results} are detailed results for the NIAH tasks shown in Tables \ref{tab:ruler_4k_results} and \ref{tab:ruler_124k_results}.
S1-S3 denote the Single NIAH 1-3 tasks, and M1-M3 denote the Multi-key NIAH 1-3 tasks.

\begin{table}[h]
\centering
\small
\begin{tabular}{lccc}
\hline
\multirow{2}{*}{Tasks} & \multicolumn{3}{c}{Single NIAH} \\
                       & \multicolumn{1}{c}{S1} & \multicolumn{1}{c}{S2} & \multicolumn{1}{c}{S3} \\
\hline
Normal+0\% & \textbf{100.0}          & 98.4           & 97.6           \\
\begin{tabular}{@{}l@{}} Normal+15\% \\ (CacheBlend) \end{tabular} & \textbf{100.0}          & \textbf{100.0}          & \textbf{100.0} \\
Normal+100\% & 100.0          & 100.0          & 100.0          \\
\hline
Block-attention+0\% & \textbf{100.0}          & 99.8           & 100.0          \\
\begin{tabular}{@{}l@{}} Block-attention+15\% \\ (Proposed) \end{tabular} & \textbf{100.0} & \textbf{100.0} & \textbf{100.0} \\
Block-attention+100\% & 100.0          & 100.0          & 100.0          \\
\hline
\end{tabular}
\caption{Evaluation results for the Single NIAH of the RULER benchmark at a 4k-token input using Llama3.1-8B-Instruct.}
\label{tab:ruler_single_niah_4k_results}
\end{table}

\begin{table}[h]
\centering
\small
\begin{tabular}{lccc}
\hline
\multirow{2}{*}{Tasks} & \multicolumn{3}{c}{Multi-key NIAH} \\
                       & \multicolumn{1}{c}{M1} & \multicolumn{1}{c}{M2} & \multicolumn{1}{c}{M3} \\
\hline
Normal+0\% & 70.2          & 88.6          & 80.0          \\
\begin{tabular}{@{}l@{}} Normal+15\% \\ (CacheBlend) \end{tabular} & 91.0          & 95.8          & 91.8 \\
Normal+100\% & 100.0         & 100.0         & 99.8          \\
\hline
Block-attention+0\% & 93.0          & \textbf{98.8} & 89.2          \\
\begin{tabular}{@{}l@{}} Block-attention+15\% \\ (Proposed) \end{tabular} & \textbf{95.8} & 98.4 & \textbf{96.6} \\
Block-attention+100\% & 100.0         & 100.0         & 99.8          \\
\hline
\end{tabular}
\caption{Evaluation results for the Multi-key NIAH of the RULER benchmark at a 4k-token input using Llama3.1-8B-Instruct.}
\label{tab:ruler_multikey_niah_4k_results}
\end{table}

\begin{table}[h]
\centering
\small
\begin{tabular}{lccc}
\hline
\multirow{2}{*}{Tasks} & \multicolumn{3}{c}{Single NIAH} \\
                       & \multicolumn{1}{c}{S1} & \multicolumn{1}{c}{S2} & \multicolumn{1}{c}{S3} \\
\hline
Normal+0\% & 0.0 & 0.0 & 0.0 \\
\begin{tabular}{@{}l@{}} Normal+15\% \\ (CacheBlend) \end{tabular} & 99.2 & 83.2 & 97.4 \\
Normal+100\% & 100.0 & 100.0 & 99.6 \\
\hline
Block-attention+0\% & 3.2 & 4.4 & 2.0 \\
\begin{tabular}{@{}l@{}} Block-attention+15\% \\ (Proposed) \end{tabular} & \textbf{100.0} & \textbf{95.0} & \textbf{99.8} \\
Block-attention+100\% & 100.0 & 100.0 & 100.0 \\
\hline
\end{tabular}
\caption{Evaluation results for the Single NIAH of the RULER benchmark at a 124k-token input using Llama3.1-8B-Instruct.}
\label{tab:ruler_single_niah_124k_results}
\end{table}

\begin{table}[h]
\centering
\small
\begin{tabular}{lccc}
\hline
\multirow{2}{*}{Tasks} & \multicolumn{3}{c}{Multi-key NIAH} \\
                       & \multicolumn{1}{c}{M1} & \multicolumn{1}{c}{M2} & \multicolumn{1}{c}{M3} \\
\hline
Normal+0\% & 0.0 & 0.0 & 0.0 \\
\begin{tabular}{@{}l@{}} Normal+15\% \\ (CacheBlend) \end{tabular} & 51.6 & 14.8 & 0.0 \\
Normal+100\% & 97.8 & 92.2 & 81.0 \\
\hline
Block-attention+0\% & 6.4 & 0.0 & 0.0 \\
\begin{tabular}{@{}l@{}} Block-attention+15\% \\ (Proposed) \end{tabular} & \textbf{73.8} & \textbf{31.4} & \textbf{2.2} \\
Block-attention+100\% & 99.0 & 92.2 & 85.0 \\
\hline
\end{tabular}
\caption{Evaluation results for the Multi-key NIAH of the RULER benchmark at a 124k-token input using Llama3.1-8B-Instruct.}
\label{tab:ruler_multikey_niah_124k_results}
\end{table}

\section{Data Proportions}
\label{sec:appendix_data}

Table \ref{tab:qunantity_fine-tune} shows the number of training samples used to train the Normal model and the Block-attention model, respectively.
Note that samples from the first part that have code-related context are used only for full attention.
As a result, the number of training samples for Block-attention is smaller than that for Normal/Block-attention.
Table \ref{tab:qunantity_qa} shows the number of evaluation samples for the four RAG benchmarks.
Tables \ref{tab:ruler_num_documents_llama} and \ref{tab:ruler_num_documents_qwen2} show the number of chunks for the RULER benchmark using Llama3.1-8B-instruct and Qwen2.5-7B-Instruct, respectively.

\begin{table}[h]
\centering
\begin{tabular}{lcc}
\hline

Dataset & \begin{tabular}{c} Normal and \\ Block-attention \end{tabular} & Block-attention \\
\hline
Tulu3 & 40,000  & 33,532  \\
2wiki & \textasciitilde 20,000 & \textasciitilde 20,000 \\
TQA   & \textasciitilde 20,000 & \textasciitilde 20,000 \\
\hline
\end{tabular}
\caption{Quantities of data used during model fine-tuning.}
\label{tab:qunantity_fine-tune}
\end{table}

\begin{table}[h]
\centering
\begin{tabular}{lc}
\hline
Dataset & Quantity \\
\hline
2wiki & 12576 \\
HQA   & 7405  \\
NQ    & 3610  \\
TQA   & 11313 \\
\hline
\end{tabular}
\caption{Quantities of Document QA data.}
\label{tab:qunantity_qa}
\end{table}

\begin{table}[h]
\centering
\small
\begin{tabular}{lcccccc}
\hline
Tasks & 4k & 8k & 16k & 32k & 64k & 124k \\
\hline
S1 & 8.0 & 16.0 & 33.0 & 65.0 & 130.0 & 252.0 \\
S2 & 7.4 & 16.1 & 32.8 & 65.8 & 132.4 & 256.8 \\
S3 & 7.6 & 15.5 & 32.3 & 65.8 & 132.6 & 256.6 \\
M1 & 7.4 & 16.1 & 32.6 & 65.8 & 132.3 & 256.8 \\
M2 & 7.9 & 15.8 & 32.6 & 64.8 & 130.0 & 252.3 \\
M3 & 8.4 & 16.4 & 33.1 & 67.7 & 137.3 & 268.0 \\
MQ & 7.4 & 15.9 & 32.8 & 66.0 & 132.6 & 256.5 \\
MV & 7.6 & 15.9 & 32.6 & 65.2 & 132.5 & 256.9 \\
VT & 8.2 & 17.0 & 33.0 & 65.0 & 130.0 & 252.0 \\
CWE & 8.0 & 16.0 & 32.0 & 64.2 & 128.7 & 249.0 \\
FWE & 7.9 & 14.8 & 29.5 & 61.6 & 124.9 & 241.8 \\
QA1 & 25.4 & 52.2 & 101.5 & 197.5 & 385.5 & 749.5 \\
QA2 & 24.3 & 55.9 & 118.4 & 243.1 & 484.4 & 944.2 \\

\hline
\end{tabular}
\caption{Average number of chunks for the RULER benchmark using Llama3.1-8B-Instruct. Note that the number of chunks for the RULER HQA is slightly different from that for QA2.}
\label{tab:ruler_num_documents_llama}
\end{table}

\begin{table}[h]
\centering
\small
\begin{tabular}{lcccccc}
\hline
Tasks & 4k & 8k & 16k & 32k & 64k & 124k \\
\hline
S1 & 8.0 & 16.0 & 33.0 & 65.0 & 130.0 & 252.0 \\
S2 & 7.8 & 15.1 & 32.8 & 65.9 & 132.7 & 256.8 \\
S3 & 7.6 & 16.0 & 32.4 & 65.6 & 132.3 & 256.6 \\
M1 & 7.3 & 16.0 & 32.5 & 65.3 & 132.3 & 256.8 \\
M2 & 7.9 & 15.8 & 32.3 & 64.8 & 130.6 & 252.8 \\
M3 & 9.0 & 17.8 & 35.3 & 68.7 & 141.4 & 277.6 \\
MQ & 7.9 & 15.8 & 32.9 & 65.8 & 132.9 & 256.4 \\
MV & 7.7 & 15.6 & 32.4 & 65.6 & 132.6 & 256.8 \\
VT & 8.2 & 17.2 & 33.2 & 65.2 & 130.2 & 252.2 \\
CWE & 8.7 & 16.8 & 32.8 & 64.9 & 129.2 & 250.0 \\
FWE & 7.9 & 14.9 & 29.6 & 61.7 & 124.8 & 241.9 \\
QA1 & 24.9 & 50.8 & 98.2 & 190.0 & 374.2 & 721.8 \\
QA2 & 23.2 & 52.7 & 112.1 & 228.9 & 459.1 & 896.7 \\

\hline
\end{tabular}
\caption{Average number of chunks for the RULER benchmark using Qwen2.5-7B-Instruct.}
\label{tab:ruler_num_documents_qwen2}
\end{table}

\section{Comparison of TTFT between Flashinfer and FlashAttention-2}
\label{sec:appendix_ttft}

\begin{figure}[t]
  \includegraphics[width=\columnwidth]{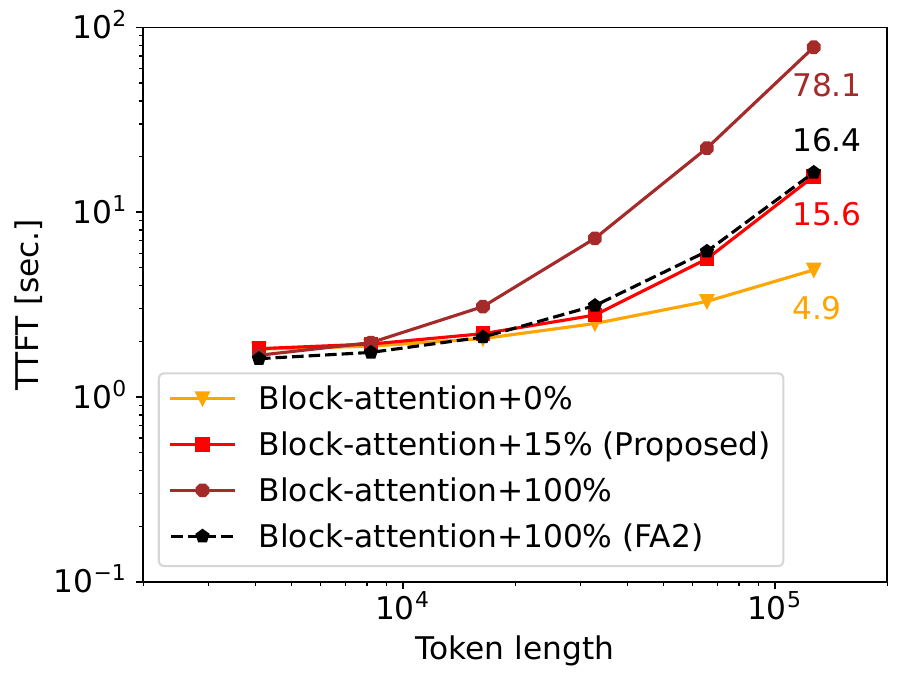}
  \caption{Averaged TTFT over 20 tasks in the RULER HQA using Llama3.1-8B-Instruct with the number of documents per cache file set to 16.
           TTFT of full attention with FlashAttention-2 (FA2) is comparable to that of 15\% recomputation with Flashinfer.}
  \label{fig:token_ttft_results_16packs_fa2}
\end{figure}

Figure \ref{fig:token_ttft_results_16packs_fa2} plots the relationship between input context length and TTFT.
In addition to the results in Fig. \ref{fig:token_ttft_results_16packs}, TTFT of full attention with FlashAttention-2 is included.
Because FlashAttention-2 reduces computation time, TTFT of full attention with FlashAttention-2 is comparable to that of 15\% recomputation with Flashinfer.

\section{Effect of the overhead incurred per file on TTFT}
\label{sec:appendix_rule_hqa_1documents_per_cache}

\begin{figure}[t]
  \includegraphics[width=\columnwidth]{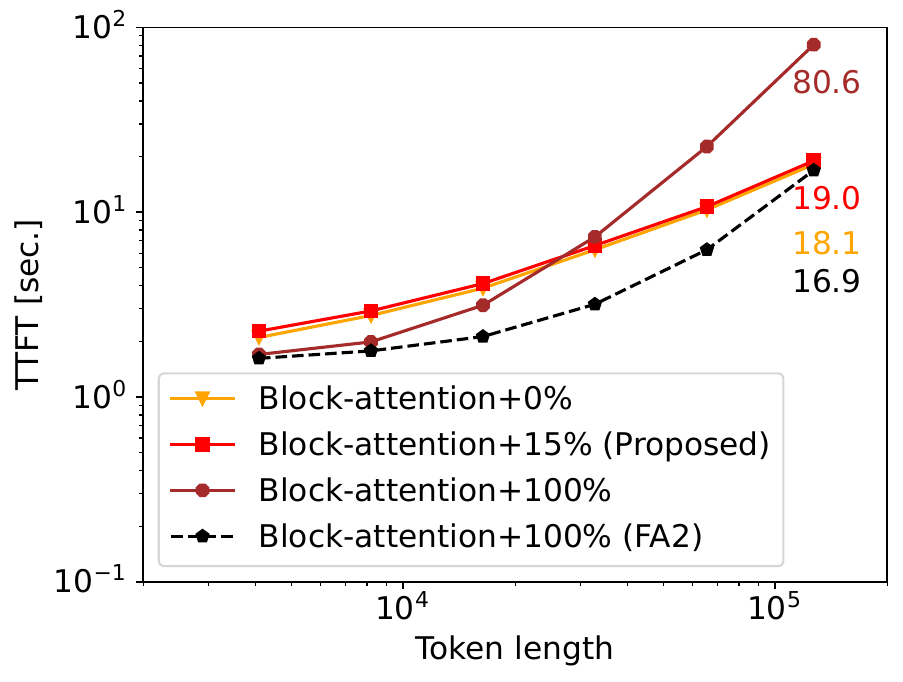}
  \caption{Averaged TTFT over 20 tasks in the RULER HQA using Llama3.1-8B-Instruct with the number of documents per cache file set to one.}
  \label{fig:token_ttft_results_1pack}
\end{figure}

Figure \ref{fig:token_ttft_results_1pack} plots the relationship between input context length and TTFT when the number of documents per cache file is one.

\begin{figure}[t]
  \includegraphics[width=\columnwidth]{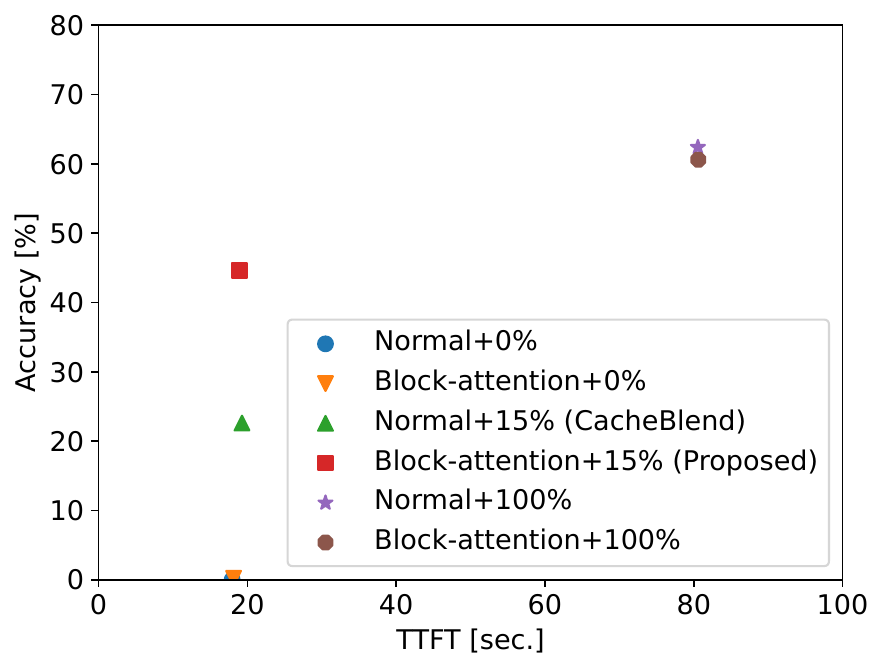}
  \caption{Relationship between averaged TTFT over 20 tasks and the accuracy in the RULER HQA at a 124k-token input using Llama3.1-8B-Instruct with the number of documents per cache file to one.}
  \label{fig:ttft_accuracy_results_1pack}
\end{figure}

TTFT of Block-attention is comparable to that of the proposed approach because a large number of KV cache files adds overhead beyond the I/O bandwidth-limited read time.
In this case, the context length needs to be 32k tokens or more to take advantage of loading KV caches. 
Figure \ref{fig:ttft_accuracy_results_1pack} shows the relationship between TTFT and the RULER HQA accuracy under this condition.

\section{RULER benchmark results using KVShare}
\label{sec:appendix_kvshare}

\begin{figure}[t]
  \includegraphics[width=\columnwidth]{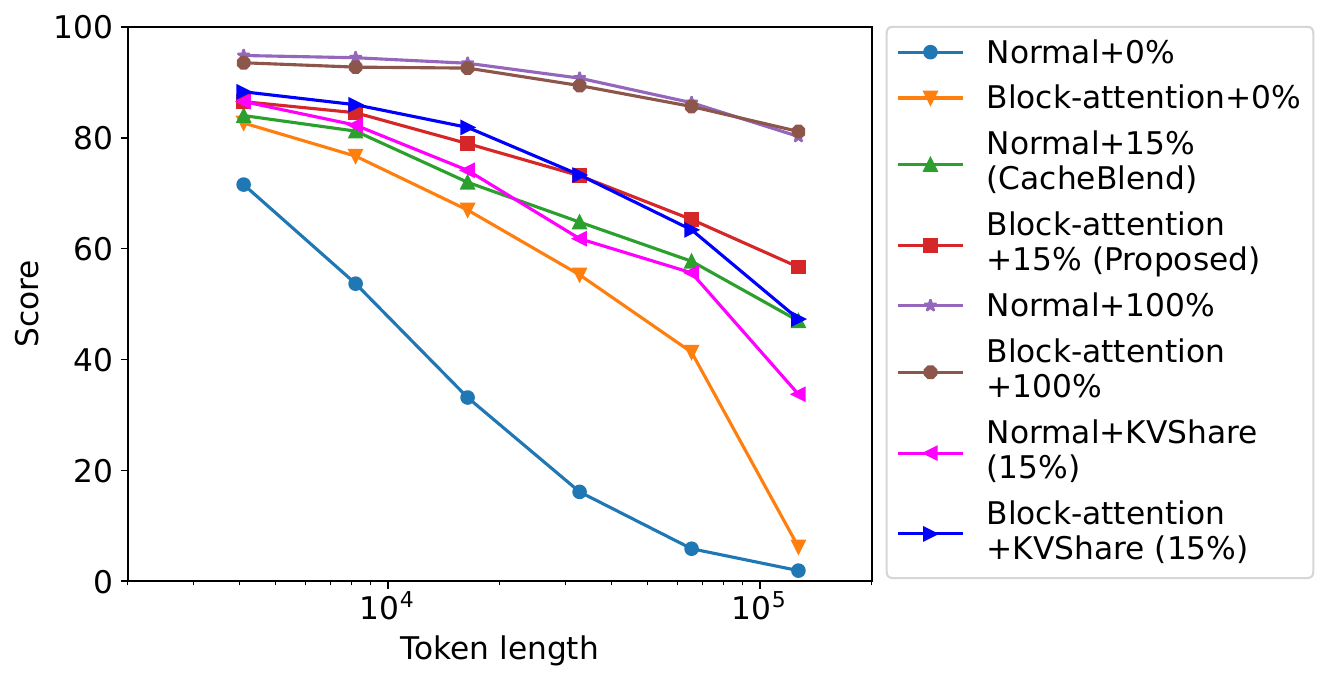}
  \caption{Evaluation results for the RULER benchmark across several input context lengths using Llama3.1-8B-Instruct. Combination of fine-tuned model and KVShare also achieves better score than KVShare alone.}
  \label{fig:ruler_kvshare_llama}
\end{figure}

\begin{figure}[t]
  \includegraphics[width=\columnwidth]{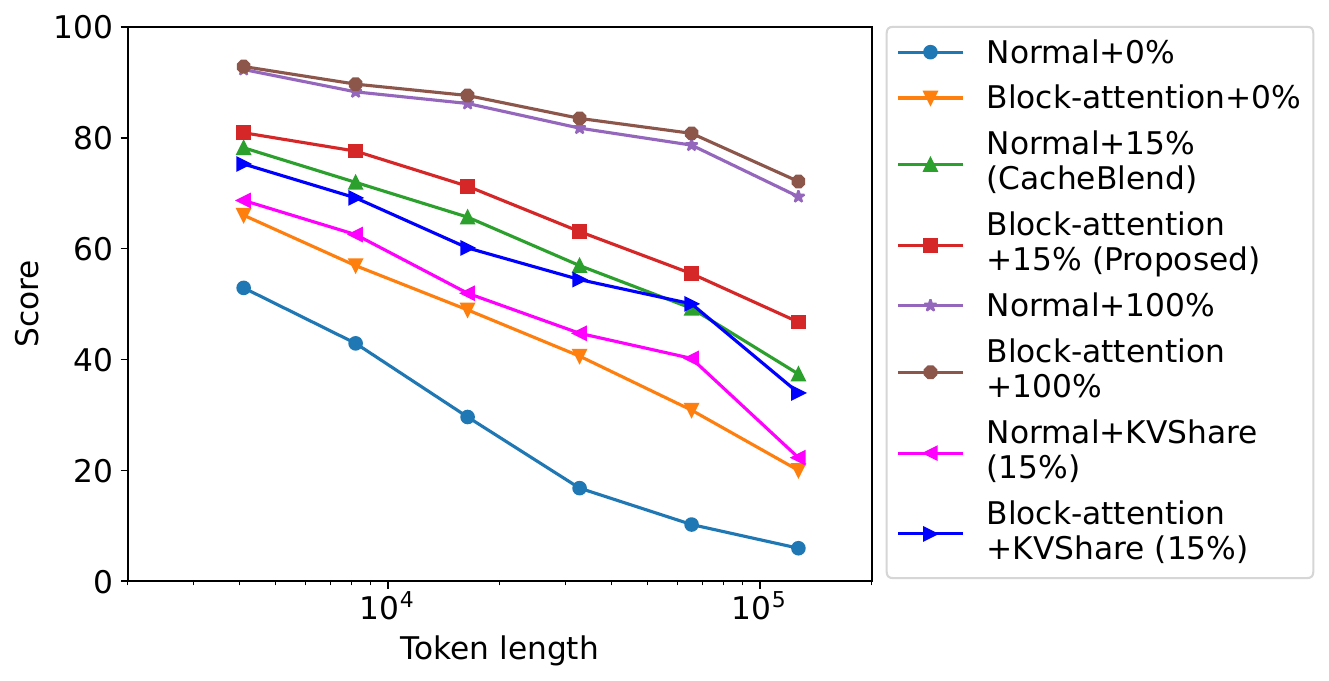}
  \caption{Evaluation results for the RULER benchmark across several input context lengths using Qwen2.5-7B-Instruct. Combination of fine-tuned model and KVShare also achieves better score than KVShare alone.}
  \label{fig:ruler_kvshare_qwen2}
\end{figure}

We evaluated KVShare, one of the attention-score-based recomputing methods. Note that KVShare was employed only in the prefill stage.
Figures \ref{fig:ruler_kvshare_llama} and \ref{fig:ruler_kvshare_qwen2} show the evaluation results for the RULER benchmark using Llama3.1-8B-Instruct and Qwen2.5-7B-Instruct, respectively.
Although the combination of fine-tuned model and recomputing method is still effective in KVShare, CacheBlend achieves better score at 124k-token input.

\section{Evaluation in the RULER HQA and four RAG benchmarks}
\label{sec:appendix_rule_hqa}

We used the input context shown below for the RULER HQA and four RAG benchmarks, as shown in Figure \ref{fig:prompt-example}.
The number of chunks in the RULER HQA in Fig. \ref{fig:ruler_hqa} is slightly different from that in QA2 in the RULER benchmark because of the slightly different handling of the QA task. 

\begin{figure}[h]
\centering
\fbox{
\begin{minipage}{0.9\linewidth}
\{special tokens for system\}
\vspace*{1\baselineskip}

You are an intelligent AI assistant. Please answer questions based on the user's instructions. Below are some reference documents that may help you in answering the user's question.
\vspace*{1\baselineskip}

\{special tokens for separation\}- Title: \{title 0\}

\{document 0\}

\{special tokens for separation\}- Title: \{title 1\}

\{document 1\}

...

\{special tokens for separation\}- Title: \{title n-1\}
\{document n-1\}

\{special tokens for separation\}
\vspace*{1\baselineskip}

Please write a high-quality answer for the given question using only the provided search documents (some of which might be irrelevant).

Question: \{question\}

\{special tokens for assistant\}
\end{minipage}
}
\caption{Prompt example used in the RULER HQA and four RAG benchmarks.}
\label{fig:prompt-example}
\end{figure}

\section{Use of AI Assistants}
\label{sec:appendix_ai_assistance}

We used AI assistants for paraphrasing support during the writing of the manuscript.
Moreover, as mentioned in Section \ref{experiments}, the generated answers for the training dataset were produced with GPT-4o mini and compiled into a dataset.

\end{document}